# Interfering Paths in Decision Trees: A Note on Deodata Predictors

**Cristian Alb**

CA.PUBLICUS@GMAIL.COM

**Abstract**
A technique for improving the prediction accuracy of decision trees is proposed. It consists in evaluating the tree's branches in parallel over multiple paths. The technique enables predictions that are more aligned with the ones generated by the nearest neighborhood variant of the deodata algorithms. The technique also enables the hybridization of the decision tree algorithm with the nearest neighborhood variant.

## 1  Introduction

Decision trees are popular and effective algorithms for prediction and classification [1]. The ID3 [2] and C4.5 [3] algorithms are well known implementations. The decision tree used as a reference in this proposal is based on them.

In [4], a set of algorithms are described that appear to achieve better prediction accuracy than decision trees. Herein, of particular interest is the variant referred to as "proximity concurrent data predictor", or "deodata delanga". The variant could also be described as "nearest neighborhood" or "all nearest neighbors". In this document the variant will be referred to as "all-NN".

The operation of the all-NN algorithm consists in searching for the set of training entries that have the largest number of query attribute matches. The resulting set of entries will be used for prediction. For instance, the relative frequency of the outcomes in the predictive set provides an indication of their likelihood. For classification, a majority rule can be applied to the predictive set.

The proposed interfering path technique endeavors to align the prediction of the decision tree with that of the all-NN algorithm.

## 2  Interfering Paths

When evaluating a decision tree, a query follows a path that starts at the root of the tree and advances towards the leaf nodes.

The interfering path technique gets activated when the node evaluation of an attribute fails. An attribute value mismatch occurs when no branch of the current node matches the attribute value of the query. In such a situation there is no matching sub-branch to follow. The default procedure stops and evaluates for prediction all outcomes associated to the problematic node.

The proposed interfering path technique forces the evaluation of paths in parallel on all the sub-branches of the problematic mismatch node. On each of the parallel branches, the remaining attributes get evaluated as in the normal procedure. The outcome subsets resulting from the evaluation of each parallel path get aggregated into a set of outcomes. The aggregated set constitutes the predictive set that will be evaluated. The predictive set is a subset of the outcomes that are normally associated to the problematic node that activated the interfering path procedure. It is a distilled selection that excludes outcomes corresponding to excessive attribute mismatches.

The procedure can be thought in analogy to the double-slit experiment in quantum physics. A particle can be detected when it passes through one slit or the other. In this analogy, the particle is the query and the slits correspond to possible attribute values. If the attribute value of the query doesn't match any of the slit attribute values, it passes simultaneously, as a wave, through all slits and ends up producing an interference pattern. The interference pattern exposes some leaf nodes and hides others.

## 3 Example

In Table 1 a toy training data set is shown. The data set will be used to illustrate the method.

| index | outcome | attributes | | |
|---|---|---|---|---|
| | | A | B | C |
| 00 | t1 | a0 | b0 | c1 |
| 01 | t2 | a0 | b0 | c1 |
| 02 | t0 | a0 | b0 | c2 |
| 03 | t0 | a0 | b0 | c2 |
| 04 | t2 | a0 | b0 | c2 |
| 05 | t0 | a1 | b0 | c0 |
| 06 | t2 | a1 | b0 | c0 |
| 07 | t1 | a1 | b0 | c2 |
| 08 | t1 | a0 | b1 | c0 |
| 09 | t0 | a2 | b1 | c0 |
| 10 | t2 | a2 | b1 | c0 |
| 11 | t0 | a0 | b1 | c1 |
| 12 | t1 | a0 | b1 | c1 |
| 13 | t2 | a1 | b1 | c1 |
| 14 | t2 | a1 | b1 | c1 |

Table 1: Classifier training data set.

A decision tree implementing a predictor for the training data set is shown in Fig. 1. Represented are the node evaluation paths followed for two queries: **X** (in red) and **Y** (in green). In normal conditions, the evaluation proceeds in the default mode; the resulting predictions are made on the basis of the outcomes associated to the corresponding leaf nodes.

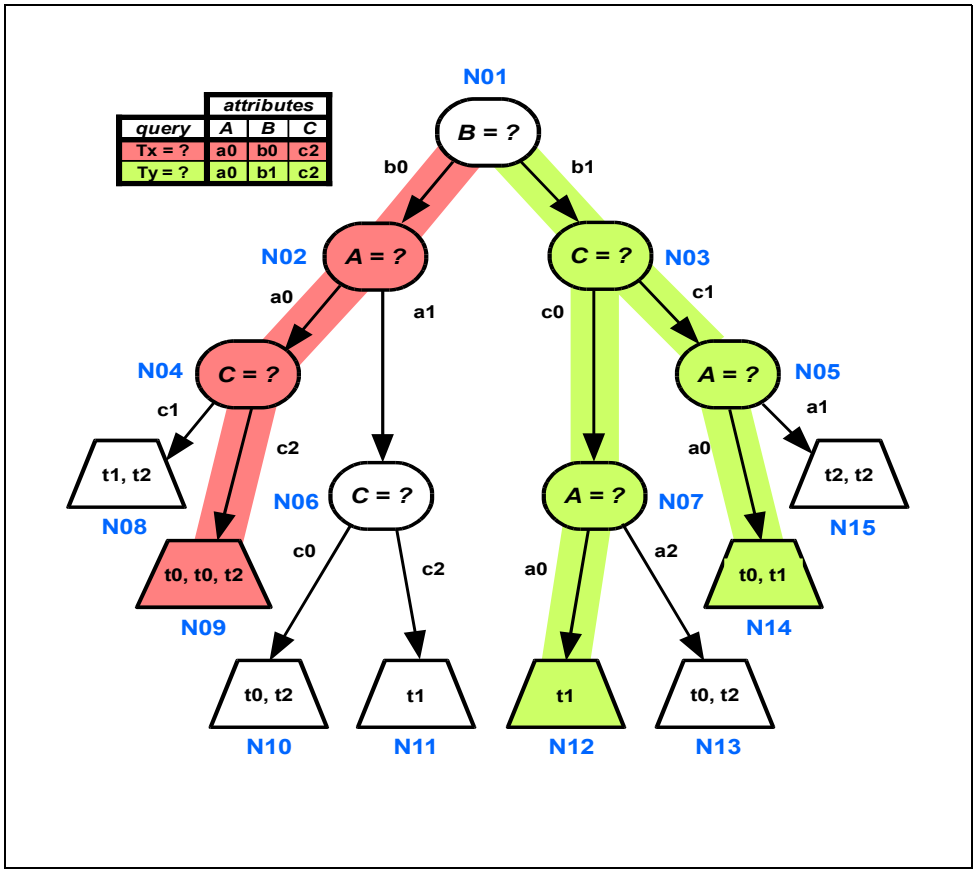

Figure 1: Classifier decision tree.

For query **X** (**A**='**a0**', **B**='**b0**', **C**='**c2**'), for each node along the path there is a sub-branch matching the corresponding attribute value of the query. The leaf node **N09** has the selection of training outcomes that match perfectly all the attributes of the query.

For query **Y** (**A**='**a0**', **B**='**b1**', **C**='**c2**'), the root node **N01** evaluates attribute **B**. There is a branch matching the query's attribute value, '**b1**'. The evaluation proceeds to node **N03**. The next evaluated attribute is **C**. The corresponding query attribute value for **C** is '**c2**'. However, node **N03** lacks a sub-branch for '**c2**'. In this situation, the standard decision tree algorithm would stop and would evaluate all outcomes associated to the node. That consists in the aggregation of leaf nodes **N12**, **N13**, **N14**, and **N15**. In that aggregation, the highest count of outcomes, three, corresponds to '**t2**'. For a classification task this would be the predicted outcome, or class label. But in the case of the interfering path technique, the evaluation is pushed forward in parallel on all the sub-branches of **N03** in order to refine the predictive set.

On branch '**c0**' the evaluation proceeds to node **N07** and the evaluated attribute is **A**. The branch matching the query attribute is '**a0**' and the leaf node **N12** is reached. The node will be selected for aggregation with the other leaf nodes of the parallel paths.

On branch '**c1**' the evaluation proceeds through node **N05** to leaf node **N14**. The final outcome predicting set will consist in the aggregation of nodes **N12** and **N14**. In this aggregation set, the highest count of outcomes, two, corresponds to '**t1**'. In a classification setting, this would be the predicted outcome.

## 4  Limitation and Extension

The above technique does provide an improvement over the regular decision tree operation; it provides a refined selection of outcomes on which to base the prediction. However, it is not a complete solution. For instance, in the previous example, the selected prediction set for query **Y** contains outcomes from leaf nodes **N12** and **N14**. All the outcomes in the set are associated to training entries that have two attributes matching query **Y**. However, the outcomes corresponding to query **X** also have two attributes that match query **Y**. If all attributes are considered equally important, those outcomes should be aggregated into the predictive set of query **Y**. In order to do that systematically, more processing is required. The resulting aggregated set would have outcome **'t0'** as the outcome with the highest count, three.

The basic technique can be extended such that it thoroughly selects the outcomes that most closely match the attributes of the query. It is conceivable that such an extended technique generates predictions equivalent to the ones provided by all-NN. This would require the decision tree to preserve information regarding the attribute values associated to the outcomes in leaf nodes. Conceptually, Fig. 2 illustrates how rows of the training table are associated to leaf nodes in order to satisfy the requirement.

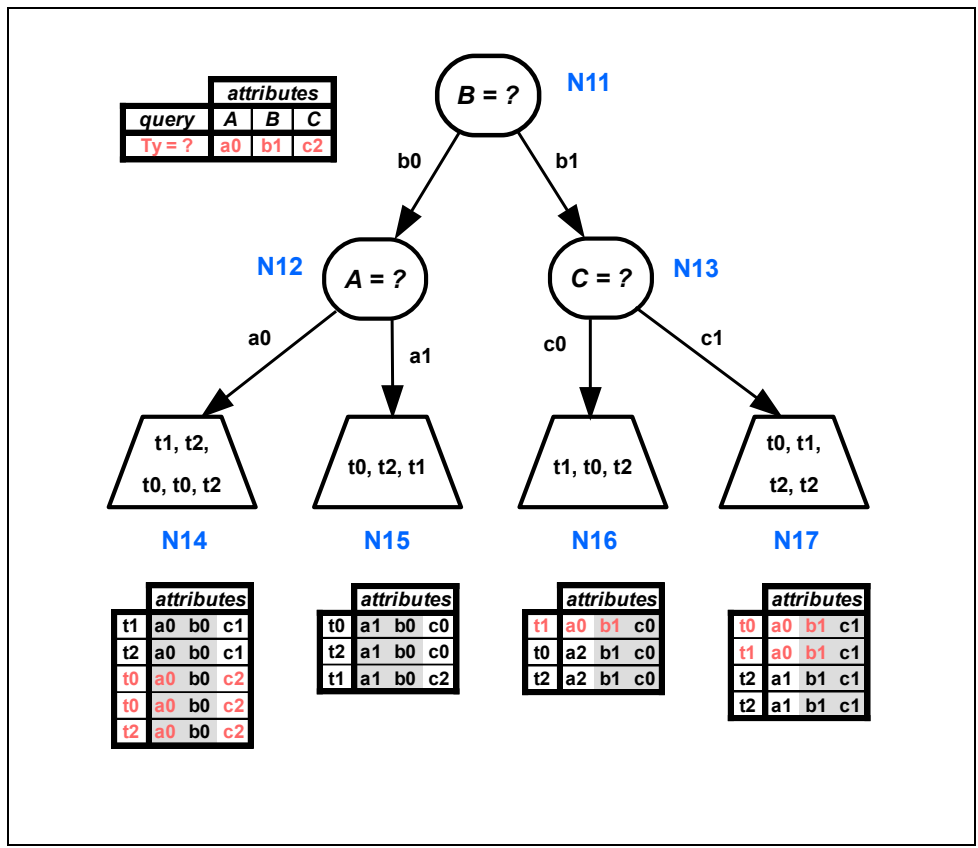

Figure 2: Hybrid decision tree.

## 5   Hybridization

What would be the use of adapting the decision tree algorithm to match the predictions of all-NN? Why not use the latter algorithm if accuracy is better? One criterion could be computational efficiency. The deodata predictors are lazy algorithms that consume processing power at the moment of the prediction. The decision tree consumes most of the processing power during the training phase.

The modified decision tree algorithm could do fast predictions in a standard way when all the attributes of a query are matched along the tree path. If there are mismatches, the more computationally intensive procedure can be used for accuracy. This allows the realization of a hybrid between decision trees and all-NN.

Computation power savings can be made by keeping track of the highest count of attribute matches. The champions are the outcome entries that reach that count. During the evaluation of a query, as soon as the number of mismatches exceeds the threshold set by the current champions, the evaluation of the current item can be abandoned. This technique is applicable also to the search for outcome entries in all-NN. However, when used with the hybrid scheme, and when a branch evaluation is abandoned, the savings are multiplied by the number of outcomes that the branch holds.

Also, outcomes in a tree branch have common attributes as shown by the shaded columns in Fig. 2. Computation savings can result from skipping the evaluation of those.

As an example, in Fig. 2 query **Y** is shown. Colored in red are the outcome entries having two attributes matching the attributes of query **Y**. Again, outcome **'t0'** has the most occurrences and, in a classification setting, would be chosen as the prediction for query **Y**.